\documentclass[silent]{article}

\usepackage[preprint,nonatbib]{nips_2018}
\usepackage{wrapfig}
\usepackage{mathtools}
\usepackage{amsmath}

\usepackage{microtype}
\usepackage{graphicx}
\usepackage{subcaption}
\usepackage{booktabs}
\usepackage{color,soul}
\usepackage{hyperref}
\usepackage{amsmath}
\usepackage{amssymb}

\usepackage[utf8]{inputenc} 
\usepackage[T1]{fontenc}    
\usepackage{hyperref}       
\usepackage{url}            
\usepackage{booktabs}       
\usepackage{amsfonts}       
\usepackage{nicefrac}       
\usepackage{microtype}      

\title{Computationally Efficient Measures of Internal Neuron Importance}

\author{
  Avanti Shrikumar*$\dagger$ \\
  Stanford University\\
  \texttt{avanti@stanford.edu} \\
  \And
  Jocelin Su* \\
  Evergreen Valley High School\\
  \texttt{jocelin.su@gmail.com} \\
  \And
  Anshul Kundaje$\dagger$ \\
  Stanford University\\
  \texttt{akundaje@stanford.edu} \\
  \And
  *co-first authors\\
  $\dagger$\textbf{ co-corresponding authors}
}

\begin{document}

\maketitle

\begin{abstract}
The challenge of assigning importance to individual neurons in a network is of interest when interpreting deep learning models. In recent work, Dhamdhere et al. proposed \emph{Total Conductance}, a "natural refinement of Integrated Gradients" for attributing importance to internal neurons. Unfortunately, the authors found that calculating conductance in tensorflow required the addition of several custom gradient operators and did not scale well. In this work, we show that the formula for Total Conductance is mathematically equivalent to Path Integrated Gradients computed on a hidden layer in the network. We provide a scalable implementation of Total Conductance using standard tensorflow gradient operators that we call Neuron Integrated Gradients. We compare Neuron Integrated Gradients to DeepLIFT, a pre-existing computationally efficient approach that is applicable to calculating internal neuron importance. We find that DeepLIFT produces strong empirical results and is faster to compute, but because it lacks the theoretical properties of Neuron Integrated Gradients, it may not always be preferred in practice.

Colab notebook reproducing results:\\ \url{http://bit.ly/neuronintegratedgradients}

\end{abstract}

\section{Introduction}
\label{sec:intro}

The challenge of explaining the predictions of neural networks is of key interest in modern machine learning. Several methods have been proposed for explaining predictions in terms of the contributions of the input \cite{shap,lime,integratedgrads,deeplift,predictiondifference,patternnet,variationaldropout,maximumentropy,l2x,maximizinginvariant,explainableai}. Though many of these methods can be applied to the task of assigning contribution scores to individual neurons, this application has not been widely explored in the literature. Recently, Leino et al.\cite{internalinfluence} and Dhamdhere et al. \cite{conductance} proposed new methods specifically aimed at computing internal neuron importance. Leino et al. presented a metric known as \emph{internal influence}, but (as shown by Dhamdhere et al. \cite{conductance}) their metric can produce poor results in practice because it does not account for the scale of the activations of the neurons.  As an alternative to \emph{internal influence}, Dhamdhere et al. introduced the concept of \emph{Total Conductance}, which they described as a modification of an existing approach called Integrated Gradients \cite{integratedgrads} for computing internal neuron importance. Unfortunately, they found that ``computing conductance in tensorflow involved adding several gradient operators and didn’t scale very well'' \cite{conductance}. This is in contrast to the original Integrated Gradients formulation, which can be computed using built-in tensorflow gradient operators and has an asymptotic runtime comparable to standard gradient computation.

In this work, we show that Total Conductance is mathematically equivalent to a generalized version of Integrated Gradients known as Path Integrated Gradients \cite{integratedgrads}. We show that Path Integrated Gradients can be implemented to have a similar runtime as standard Integrated Gradients and relies on built-in gradient operators, thereby constituting an approach to computing neuron importance that is scalable and easily adopted in practice. We refer to our formulation as Neuron Integrated Gradients. We compare Neuron Integrated Gradients to DeepLIFT \cite{deeplift}, a pre-existing attribution method that can be applied to the internal neurons in a network but which was not compared to in \cite{conductance}. We find that DeepLIFT empirically produces competitive results and is faster to compute than Neuron Integrated Gradients, but we also note that DeepLIFT lacks some of the theoretical guarantees of Integrated Gradients and thus may not always be preferred in practice.

\section{Contributions}
\label{sec:contributions}

Our contributions are as follows:

\begin{itemize}
\item We show a mathematical equivalence between Total Conductance and Path Integrated Gradients on neurons in a hidden layer (Sec. ~\ref{sec:equivalence}).
\item We provide a practical implementation of Total Conductance (computed with Path Integrated Gradients) using standard tensorflow gradient operators, which we refer to as \emph{Neuron Integrated Gradients} (Sec. ~\ref{sec:implementation}).
\item We compare Neuron Integrated Gradients to DeepLIFT, a pre-existing approach for computing importance scores that can be applied to internal neurons in a network (Sec. ~\ref{sec:results}).

\end{itemize}

\section{Equivalence Between Total Conductance and Path Integrated Gradients}
\label{sec:equivalence}

Path Integrated Gradients, proposed by Sundararajan et al. in \cite{integratedgrads}, is a method for explaining individual predictions of a deep learning model by assigning contribution scores to the input variables. For a given example, Path Integrated Gradients works by selecting a ``baseline'' or ``reference'' input and following a path that slowly transforms the ``reference'' input to the actual value of the input. At each step of the path, the gradient on the input is computed and multiplied by the step taken. The sum of ``$\text{gradient }\times\text{ step}$'' over all the steps is equal to the contribution scores on the input variables. In the limit of infinitesimal steps, we get the formula:

\begin{equation}
\label{eqn:pathintgrad}
\text{PathIntegratedGrads}_i^\gamma \text{ ::= } \int_{\alpha=0}^{1} \frac{\partial F(\gamma(\alpha))}{\partial \gamma_i(\alpha)}\frac{\partial \gamma_i (\alpha)}{\partial \alpha} d\alpha 
\end{equation}

Where:
\begin{itemize}
\item $\gamma : [0,1] \rightarrow \mathbb{R}^n$ is a smooth function specifying a path in $\mathbb{R}^n$ from the reference input $x'$ to the actual input $x$
\item $n$ is the dimensionality of $x$
\item $\gamma_i(\alpha)$ refers to the value of the $i$th element of $\gamma(\alpha)$
\item $\frac{\partial \gamma_i (\alpha)}{\partial \alpha} d\alpha$ is an infinitesimal step along the path for element $i$
\end{itemize}
In the original paper, $x$ was taken to refer to the input of the first layer of the network. However, as we will show, it can just as easily be taken to refer to any intermediate layer of the network.

For the straightline path $\gamma(\alpha) = x' + \alpha(x - x')$, the authors show that one obtains the formula for conventional Integrated Gradients (equation 1 in the integrated gradients paper), reproduced below:

\begin{equation}
\text{IntegratedGrads}_i(x) \text{ ::= } (x_i - x_i') \times \int_{\alpha=0}^{1} \frac{\partial F(x' + \alpha(x - x'))}{\partial x_i} d\alpha
\end{equation}

Note that the formula above (taken verbatim from the original paper) overloads notation for $x_i$. The $x_i$ in the partial derivative $\frac{\partial F(x' + \alpha(x - x'))}{\partial x_i}$ is equivalent to $\gamma_i(\alpha)$ in the formula for PathIntegratedGrads, and is thus a function of $\alpha$. Specifically, $\gamma_i(\alpha) = x_i' + \alpha(x_i - x_i')$. However, the $x_i$ in the $(x_i - x_i')$ term, which lies outside the integral, comes from $\frac{\partial \gamma_i(\alpha)}{\partial \alpha} = \frac{\partial(x_i' + \alpha(x_i - x_i'))}{\partial \alpha}$, and thus refers to the final value of input $i$ (which is not a function of $\alpha$). To avoid confusion, we will denote the $x_i$ in the denominator as $\gamma_{x_i}(\alpha)$, giving us the formula:

\begin{equation}
\text{IntegratedGrads}_i(x) \text{ ::= } (x_i - x_i') \times \int_{\alpha=0}^{1} \frac{\partial F(x' + \alpha(x - x'))}{\partial \gamma_{x_i}(\alpha)} d\alpha
\end{equation}

We now revisit the formula for the Total Conductance of a neuron $y$ (equation 3 in \cite{conductance}), reproduced verbatim below:

\begin{equation}
\text{Cond}^y(x) \text{ ::= } \sum_i (x_i - x_i') \cdot \int_{\alpha = 0}^{\alpha = 1} \frac{\partial F(x' + \alpha(x - x'))}{\partial y} \cdot \frac{\partial y}{\partial x_i} d\alpha
\end{equation}

Conductance satisfies the $\emph{completeness}$ property, in that the total conductance for any single hidden layer adds up to the difference in the predictions $F(x) - F(x')$. The authors state that this follows from the fact that Integrated Gradients satisfies completeness and  the derivative $\frac{\partial F}{\partial x_i}$ can be expressed using the chain rule as $\sum_j \frac{\partial F}{\partial y_j}\frac{\partial y_j}{\partial x_i}$, where $j$ indexes the neurons in a layer. As mentioned before, $x_i$ is a function of $\alpha$ here. For clarity, we can replace it with the notation $\gamma_{x_i}(\alpha)$. Thus, it follows that $y_j$ is a function of $\alpha$ - specifically, $y_j$ refers to the activation of the neuron $y_j$ given the input values corresponding to $\alpha$. Rewriting the formula for conductance to explicitly reveal which values are functions of $\alpha$, we have:

\begin{equation}
\text{Cond}^y(x) \text{ ::= } \sum_i (x_i - x_i') \cdot \int_{\alpha = 0}^{\alpha = 1} \frac{\partial F(x' + \alpha(x - x'))}{\partial \gamma_y(\alpha)} \cdot \frac{\partial \gamma_y(\alpha)}{\partial \gamma_{x_i}(\alpha)} d\alpha
\end{equation}

We are now in a position to show the equivalence between Conductance and Path Integrated Gradients. Recall that $\gamma_{x_i}(\alpha) = x_i' + \alpha(x_i - x_i')$. Thus, $(x_i - x_i') = \frac{\partial \gamma_{x_i}(\alpha)}{\partial \alpha}$ for all values of $\alpha$. We can therefore rewrite conductance as:

\begin{align*}
\text{Cond}^y(x) &= \sum_i \frac{\partial \gamma_{x_i}(\alpha)}{\partial \alpha} \int_{\alpha = 0}^{\alpha = 1} \frac{\partial F(x' + \alpha(x - x'))}{\partial \gamma_y(\alpha)} \cdot \frac{\partial \gamma_y(\alpha)}{\partial \gamma_{x_i}(\alpha)} d\alpha\\
&\text{Moving the summation inside the integral:}\\
&= \int_{\alpha = 0}^{\alpha = 1} \sum_i \frac{\partial \gamma_{x_i}(\alpha)}{\partial \alpha} \cdot \frac{\partial F(x' + \alpha(x - x'))}{\partial \gamma_y(\alpha)} \cdot \frac{\partial \gamma_y(\alpha)}{\partial \gamma_{x_i}(\alpha)} d\alpha\\
&\text{Reordering the terms:}\\
&= \int_{\alpha = 0}^{\alpha = 1} \frac{\partial F(x' + \alpha(x - x'))}{\partial \gamma_y(\alpha)} \cdot \left(\sum_i \frac{\partial \gamma_{x_i}(\alpha)}{\partial \alpha} \cdot \frac{\partial \gamma_y(\alpha)}{\partial \gamma_{x_i}(\alpha)} \right) d\alpha\\
&\text{Applying the chain rule:}\\
&= \int_{\alpha = 0}^{\alpha = 1} \frac{\partial F(x' + \alpha(x - x'))}{\partial \gamma_y(\alpha)} \frac{\partial \gamma_y(\alpha)}{\partial \alpha} d\alpha \addtocounter{equation}{1}\tag{\theequation} \label{eqn:equivalence}
\end{align*}
We note that Eqn.~\ref{eqn:equivalence} is equivalent to the formula for Path Integrated Gradients in Eqn.~\ref{eqn:pathintgrad} when $\gamma(\alpha)$ is a vector of activations of the entire layer containing $y$ given the input $(x' + \alpha(x - x'))$ and $\gamma_y(\alpha)$ is the activation of the specific neuron $y$ within $\gamma(\alpha)$. Because we have eliminated the summation over all inputs $i$, the Total Conductance expressed in this way can be computed in a similar runtime to standard integrated gradients and can be computed using built-in gradient operators in tensorflow. We refer to this formulation as \emph{Neuron Integrated Gradients}.

\section{Implementation of Neuron Integrated Gradients}
\label{sec:implementation}

In practice, we use a Riemann approximation to perform Neuron Integrated Gradients. Let $x^{(i)}$ be the $i^{th}$ point of an $n$-point linear interpolation of $x'$ to $x$ and let $F_y(x)$ be the activation of neuron $y$ given input $x$. We can express $x^{(i)}$ as $x' + \frac{i}{n} (x - x').$

From Eqn.~\ref{eqn:equivalence}, note that the partial derivative $\frac{\partial F(x' + \alpha(x - x'))}{\partial \gamma_y(\alpha)}$ at $\alpha = \frac{i}{n}$ is equal to $\frac{\partial F(x^{(i)})}{\partial F_y(x^{(i)})}$, and the term $\frac{\partial \gamma_y(\alpha)}{\partial \alpha} d\alpha$ (intuitively representing a ``step'' along the path) is approximately $\Delta F_y(x^{(i)}) = F_y(x^{(i)}) - F_y(x^{(i-1)})$ when $\alpha = \frac{i}{n}$ . Thus, the Riemann sum is:
\begin{equation}
\sum_{i=1}^n \frac{\partial F(x^{(i)})}{\partial y} (F_y(x^{(i)}) - F_y(x^{(i-1)}))
\end{equation}

In code, the partial derivative can be computed using a gradients operator such as {\tt tensorflow.gradients}. Like standard Integrated Gradients, the runtime is linear in the number of steps $n$ and linear in the time taken to compute $\frac{\partial F(x_i)}{\partial y}$.

\section{Application of DeepLIFT to Computing Internal Neuron Importance}

DeepLIFT \cite{deeplift} is a pre-existing method for computing contribution scores to explain individual predictions of a neural network. Like Integrated Gradients, DeepLIFT compares the behavior of the network on the actual input to its behavior on a ``reference'' input and assigns contribution scores such that the total contribution across all input variables is equal to the difference in the output from the reference output (i.e. DeepLIFT satisfies \emph{completeness}). Because DeepLIFT does not rely on computing a numerical integral, it has a faster runtime than Integrated Gradients. However, the tradeoff is that DeepLIFT is not implementation invariant; DeepLIFT uses knowledge of the internal structure of the network to decide how to assign importance scores, and could therefore give slightly different results for functionally equivalent networks.

For assigning contributions in the presence of nonlinearities, DeepLIFT suggests two heuristics: the Rescale rule and the RevealCancel rule. Ancona et al. \cite{anconadeepexplain} found in empirical evaluations that ``Integrated Gradients and DeepLIFT [with the Rescale rule] have very high correlation, suggesting that the latter is a good (and faster) approximation of the former in practice". The RevealCancel rule was added to deal with some failure cases that affect both the Rescale rule and Integrated Gradients, but it comes with its own complexities \cite{deeplift}. The default setting as of v0.6.6 of DeepLIFT is to use the Rescale rule on convolutional layers and the RevealCancel rule on fully-connected layers.

In this work, we applied DeepLIFT to assign contributions to neurons in the hidden layers of a network. We found that DeepLIFT with the default settings produced the best empirical results and was faster to compute that Neuron Integrated Gradients. However, because DeepLIFT is not implementation invariant, it may not always be preferred in practice.

\section{Results}
\label{sec:results}

We evaluated the different neuron importance scoring methods using a modification of the ablation studies in Dhamdhere et al. \cite{conductance}. For a given input, we ``ablate'' a neuron by clamping the activation of the neuron to its activation under the reference input. We measure the quality of a contribution scoring method by assessing how well the contribution scores on the neurons can serve as a proxy for the impact on the output of ablating those neurons. DeepLIFT and Integrated Gradients both satisfy the conservation property, in that the total contribution of all the neurons is equal to the impact of ablating all the neurons. However, this equality does not hold when only a subset of neurons are ablated; the degree to which it holds will serve as our quality metric.

\begin{figure}
\centering
\begin{tabular}[c]{cc}
\begin{subfigure}[b]{0.5\textwidth}
\includegraphics[width=\linewidth]{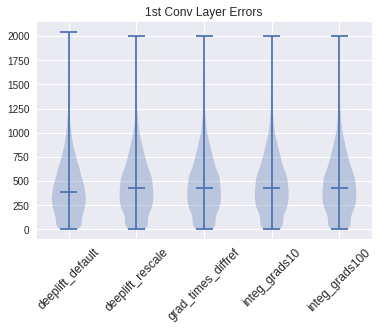}
\end{subfigure}
\begin{subfigure}[b]{0.5\textwidth}
\includegraphics[width=\linewidth]{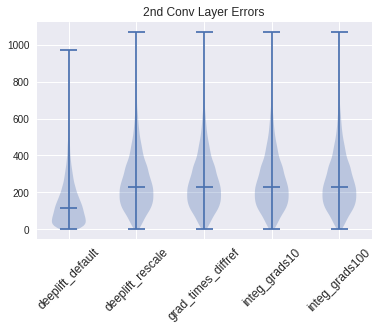}
\end{subfigure}
\end{tabular}
\caption{ In the first convolutional layer, Neuron Integrated Gradients achieved a Mean Absolute Error (MAE) of 457.89975 at $n=10$ and 458.5485 at $n=100$ ($n$ is the number of steps). DeepLIFT with default settings obtained a slightly better MAE of 422.164 (p < e-15 via a Mann-Whitney rank sum test), while DeepLIFT with only the rescale rule achieved a MAE of 454.3863 and gradient $\times$ difference-from-reference achieved 457.2978. In the second convolutional layer, Neuron Integrated Gradients achieved a Mean Absolute Error (MAE) of 255.53603 at $n=10$ and 255.7105 at $n=100$. DeepLIFT with default settings performed noticeably better with a MAE of 145.73521 (p < e-19 via a Mann-Whitney rank sum test), while DeepLIFT with only the rescale rule achieved a MAE of 255.1436 and gradient $\times$ difference-from-reference achieved 255.32726.}
\label{fig:mae}
\end{figure}

In this study, we used the MNIST classification model trained in \cite{deeplift} and computed neuron contribution scores of the first and second convolutional layers to the softmax layer (pre-nonlinearity). The reason we did not use the post-nonlinearity outputs is to avoid saturation effects, as explained in Section 3.6 of \cite{deeplift}. To account for interdependence between classes, the mean contribution across all classes was subtracted from the contribution to each single class as per eqn. 5 in \cite{deeplift} (the motivation being that if a particular neuron has the same contribution to all the softmax neurons pre-nonlinearity, it is effectively contributing to none of the softmax classes because the softmax post-nonlinearity is normalized to sum to 1). The reference used was an all-black image. For each layer, we identified the 10\% of neurons with the highest difference from their reference activation and ablated the neurons by setting them to their reference activation. Our output of interest was the value of the top softmax class (pre-nonlinearity) minus the mean output across all classes (the normalization by the mean is done to account for interdependence between classes, as described earlier). We looked at the value of this output before and after ablation and compared it to the total contribution scores of the ablated neurons. The "error" was taken to be the absolute difference between the actual change in the output and the predicted change based on the sum of the contribution scores.

\begin{figure}[h!]
\begin{center}
\centerline{\includegraphics[width=0.8\textwidth]{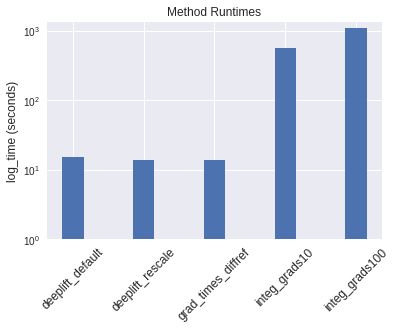}}
\caption{Neuron Integrated Gradients had a runtime of 564.12058s at $n=10$ and 1082.58269s at $n=100$. DeepLIFT with default settings, DeepLIFT with only rescale rule, and gradient $\times$ difference-from-reference (all implemented through DeepLIFT v0.6.6) ran in  15.29601, 13.67815, and 13.76395 seconds respectively. We note that the runtime of Neuron Integrated Gradients could be improved by more efficient batching; in the current implementation, each batch corresponds to the interpolation points used to interpret a single image. All runtimes were measured on Google Colab.}
\label{fig:runtime}
\end{center}
\end{figure}

We compared Neuron Integrated Gradients to DeepLIFT v0.6.6 with default settings (i.e. using the RevealCancel rule on the fully-connected layers and the Rescale rule on the convolutional layers) and DeepLIFT using the Rescale rule on all layers. Consistent with the results in Ancona et al., we found that Integrated Gradients and DeepLIFT with the Rescale rule performed very similarly. We also added a comparison to solely using the gradient on the neurons multiplied by the difference from their reference activation. Although ``gradient $\times$ difference-from-reference'' does not satisfy the conservation property for nonlinear models, it can work well as a linear approximation in some situations. Note, however, that the similar error of ``gradient $\times$ difference-from-reference'' and integrated gradients in this evaluation should not be presumed to generalize to other situations; in most cases, we do not recommend using ``gradient $\times$ difference-from-reference'' as it is susceptible to saturation effects \cite{deeplift,integratedgrads,conductance}. The results for the mean absolute errors are in Fig. ~\ref{fig:mae}, and the results for runtimes are in Fig. ~\ref{fig:runtime}.

A Colab notebook reproducing the results is at \url{http://bit.ly/neuronintegratedgradients}

\section{Conclusion}

We have presented Neuron Integrated Gradients, an efficient approach to computing the contribution scores of internal neurons in a network. Neuron Integrated Gradients inherits the benefits of standard Integrated Gradients, namely implementation invariance, and is mathematically equivalent to Total Conductance while being scalable and easy to implement. We compare Neuron Integrated Gradients to DeepLIFT, a pre-existing approach to computing importance scores that can be applied to individual neurons, and find that DeepLIFT gives strong empirical results while being faster to run. However, we note that because DeepLIFT lacks the theoretical guarantee of implementation invariance, it may not always be preferred over Neuron Integrated Gradients in practice. A Colab notebook reproducing the results is at \url{http://bit.ly/neuronintegratedgradients}.

\section{Author Contributions}

AS derived the equivalence between Total Conductance and Path Integrated Gradients. AS and JS implemented Neuron Integrated Gradients and conducted comparison experiments. JS prepared the Colab notebook. AK provided guidance and feedback. AS, JS, and AK wrote the manuscript.

\bibliography{example_paper}
\bibliographystyle{plain}

\end{document}